\documentclass[letterpaper, 10 pt, conference]{ieeeconf}  

\IEEEoverridecommandlockouts                              

\usepackage[left=0.667in,right=0.667in,top=0.792in,bottom=0.65in]{geometry}
\usepackage{moreverb,url}
\usepackage{times}
\usepackage{multicol}
\usepackage{amsmath,amsfonts,amssymb,mathtools,leftidx}
\usepackage{wrapfig}
\usepackage{float}
\usepackage{xcolor}
\usepackage{graphicx}
\usepackage{multirow}
\usepackage{mathrsfs}
\usepackage{nccmath}
\usepackage{algorithm,algpseudocode}
\usepackage{balance}
\usepackage{afterpage}
\newtheorem{mydef}{Definition}
\usepackage[textwidth=10mm]{todonotes}
\usepackage{tabularx}
\usepackage[noadjust]{cite}
\usepackage[colorlinks,bookmarksopen,bookmarksnumbered,citecolor=red,urlcolor=red]{hyperref}

\title{\LARGE \bf
Online Exploration and Coverage Planning \\ in Unknown Obstacle-Cluttered Environments}

\author{Xinyue Kan, Hanzhe Teng, and Konstantinos Karydis
\thanks{The authors are with the Dept. of Electrical and Computer Engineering, University of California, Riverside. 
	Email: {\{xkan001, hteng007, karydis\}@ucr.edu}. 
We gratefully acknowledge the support of NSF under grants \#IIS-1724341 and \#IIS-1901379, and ONR under grant \#N00014-18-1-2252.  
Any opinions, findings, and conclusions or recommendations expressed in this material are those of the authors and do not necessarily reflect the views of the funding agencies.}
}

\begin{document}

\maketitle
\thispagestyle{empty}
\pagestyle{empty}

\begin{abstract}
Online coverage planning can be useful in applications like field monitoring and search and rescue. Without prior information of the environment, achieving resolution-complete coverage considering the non-holonomic mobility constraints in commonly-used vehicles (e.g., wheeled robots) remains a challenge. In this paper, we propose a hierarchical, hex-decomposition-based coverage planning algorithm for unknown, obstacle-cluttered environments. The proposed approach ensures resolution-complete coverage, can be tuned to achieve fast exploration, and plans smooth paths for Dubins vehicles to follow at constant velocity in real-time. Gazebo simulations and hardware experiments with a non-holonomic wheeled robot show that \textcolor{black}{our approach can successfully tradeoff between coverage and exploration speed and can outperform existing online coverage algorithms in terms of total covered area or exploration speed according to how it is tuned.}
\end{abstract}

\section{Introduction}

The paper addresses \textit{online coverage planning} in unknown environments for vehicles with non-holonomic constraints. Efficient field coverage is essential for tasks such as environmental monitoring~\cite{Popovic2017}, map reconstruction~\cite{dong2017}, locating survivors~\cite{Hayat2017}, and autonomous exploration of forested areas~\cite{Anttoni2017}. In all these applications, regions to be explored may be unknown and partially observable. Even if the environment map can be obtained prior to departure, unexpected unvisitable areas may occur, such as collapsed trees following a storm. Hence, it is necessary to develop approaches that enable online exploration and coverage planning of irregularly-shaped environments with potential unexpected obstacles.  

Depending on the application, various different types of unmanned vehicles--including aerial (fixed-wing aircraft), surface, ground (wheeled/legged robots), and underwater ones (e.g.,~\cite{Shima2007,Karapetyan2018,Tokekar2014,Li2016,Karapetyan2019,Rahman2018})--can be deployed. A common challenge among most of them is the presence of non-holonomic mobility constraints, often manifested as a minimum turning curvature constraint. A way to take into consideration this constraint is by using a Dubins vehicle model~\cite{Dubins_1957}, which specifies the vehicle to move in fixed-speed straight lines and counter/clockwise turns. More complex paths can be designed by concatenating straight-line and turning maneuvers.


When the environment is known, existing approaches (e.g.,~\cite{Galceran2012,Huang2001,Lewis2017,Karapetyan2018}) decompose the region into a set of non-overlapping subregions, and then plan paths within each subregion. In practice, however, regions to be explored can be unknown. Exhaustive search strategies, like back-and-forth parallel swath motions~\cite{acar2002,Zelinsky93} or spiral paths~\cite{Bosse2007}, alter the robot's direction of motion if obstacles are encountered. This may lead to incomplete coverage when the region occluded by the obstacle has not been visited. In contrast, many existing online coverage methods~\cite{Lee2011,acar2002,Gabriely2001,Hoang2013} lead to abrupt velocity and orientation changes when encountering obstacles. This effect becomes pronounced especially when operating in obstacle-cluttered environments, \textcolor{black}{and can hinder tasks for which the success rate} is sensitive to the quality of sensor input, e.g., in field reconstruction~\cite{dong2017,Chen2019} and survivor localization~\cite{Doulamis2017}. 
To mitigate this challenge, we propose an online approach that plans smooth trajectories that minimize the frequency of acceleration-deceleration events.

To represent the environment we consider a uniform hexagonal grid where a cell's dimension is determined by the robot's sensor footprint. \textcolor{black}{Hexagon-based partitioning} enjoys several benefits, including regular tessellation~\cite{gullberg1997mathematics}, uniform travel distance to all adjacent cells, and better description of non-convex regions~\cite{algfoor2015comprehensive}.
The effectiveness of hexagonal cell decomposition has been shown in applications including potential-field-based path finding~\cite{Hou1991}, field search~\cite{Azpurua2018} and offline path 
finding~\cite{yu2018} in known environments, and online underwater mine countermeasure~\cite{paull2013} with no restricted areas. Different from those aforementioned approaches, we focus on describing \textbf{unknown}, \textbf{obstacle-cluttered}\textcolor{black}{, bounded environments with duplicates of regular hexagons}, which fill a plane with no gap or overlap.

We propose an online, hierarchical coverage planning approach for Dubins vehicles. At the high level, a Hex-Decomposition Coverage Planning (HDCP) algorithm is proposed. Based on information collected from the robot's observations \textcolor{black}{up to the current time}, the robot selects a feasible hexagon subregion to explore next. At the low level, Dubins-curve-based paths are planned in real-time. Closed-form solutions for feasible paths (e.g., start and goal positions for line segments, angles for arcs) are provided. The proposed HDCP algorithm aims to cover the entire unknown \textcolor{black}{(yet bounded)} environment, whereas its variant, HDCP-E, is used for fast exploration. The proposed method is evaluated in Gazebo simulations in three forest/farm-like environments \textcolor{black}{and in one baseline empty environment}, and compared against Spanning Tree Coverage (STC)~\cite{Gabriely2001}, Boustrophedon motions and the A* search algorithm (BA${}^*$)~\cite{Hoang2013}, and Multi-robot Hex Decomposition Exploration (M-HDE)~\cite{Kan2019} in terms of covered area and exploration speed. We observe that unknown environments containing random obstacles can be fully covered. Moreover, HDCP can cover the most amount of free space, whereas HDCP-E can achieve the highest coverage area per unit time. The proposed method is also evaluated in hardware with a non-holonomic wheeled robot.
%

{\bf Contributions:}
This paper has three key contributions. 
\begin{enumerate}
\item We develop a hex-decomposition-based online coverage planning algorithm, HDCP, that guarantees resolution-completeness in unknown, cluttered spaces.
\item We propose the variant HDCP-E, to trade-off between fast exploration and resolution-completeness.
\item We offer closed-form solutions for planning smooth paths that robots can follow at constant speed.
\end{enumerate}

\textcolor{black}{The major difference between HDCP and our previous work M-HDE~\cite{Kan2019}--which also applies hexagonal cell decomposition in unknown environments--is the sensor-footprint-based decomposition strategy herein which ensures full coverage within each subregion. 
Further, M-HDE is developed mainly for online exploration tasks. If applied to coverage tasks (as in this work) it performs worse because 
it prioritizes visiting the most unexplored area. Doing so leaves uncovered subregions and necessitates returning back to fill in holes.}


\vspace{-3pt}
\section{Related Work}
Several methods have been proposed to tackle the coverage path planning problem. When prior map information is available, planning can be offline~\cite{choset2001coverage}. 
Most planners use some form of decomposition, like Boustrophedon~\cite{Mannadiar2010,Karapetyan2018},  Semi-boustrophedon~\cite{Lewis2017}, Morse~\cite{Galceran2012}, or Line-sweep-based~\cite{Huang2001} decomposition, to partition the free space into a set of non-overlapping cells. For online coverage planning, using information collected by on-board sensors, similar cellular-decomposition-based strategies~\cite{Hoang2013,Hazon2006,acar2002} are also applied. Resolution-complete coverage can be obtained by ensuring that all cells can be visited, and then applying ``lawnmower" motions within each cell. Another popular approach used for coverage planning is the Spanning Tree algorithm~\cite{Gabriely2001,Agmon2006}. However, paths generated by those methods may contain sharp turns which can reduce efficiency and increase fuel consumption for non-holonomic robots~\cite{Khan2017OnCC}.
%
%

A common way to consider non-holonomic constraints is to generate feasible paths using a Dubins model for offline coverage planning~\cite{Xu2014,Karapetyan2018,Savla2007,Lewis2017}. Lewis et al.~\cite{Lewis2017} solve the offline coverage problem as a traveling salesman problem, and add constraints to ensure planned paths consist of line segments and curves of a given minimum radius only. Yu et al.~\cite{Yu2015} proposed a graphical-optimization-based smooth planning strategy for Dubins vehicles. The method reduces the total coverage time, but at the expense of high computational complexity. Function-based smooth coverage planning methods generate paths represented by functions like clothoids~\cite{SABELHAUS2013399} and B\'{e}zier curves~\cite{Khan2016,LEE2011801}. Due to their smoothness, B\'{e}zier curves enable fast coverage and energy efficiency, but at the price of complex calculations.

Online coverage planning methods that are directly applicable to Dubins vehicles remain limited. One approach is to obtain an offline solution for coverage paths using any available prior knowledge, and then replan according to the information collected through sensors as the robot moves to avoid collisions
~\cite{Carreras2016, Lacroix2017,Marek2019,Guo2006}. Another way is to modify existing online coverage approaches, such as online STC~\cite{Gabriely2001} and online BA${}^*$~\cite{Hoang2013}, to decelerate and make smooth turns that satisfy the minimum turning radius constraints. Our proposed work fills in the gap by utilizing sensor-based decomposition and directly incorporating non-holonomic constraints.

\section{Problem Setup}
\textcolor{black}{Consider a robot tasked to survey an unknown, bounded, obstacle-cluttered space $\mathcal{S}$.} The robot is equipped with \textit{navigation sensors} (e.g., LIDAR, depth camera) and \textit{observation sensors} (e.g., RGB/thermal camera, mine detector). Navigation sensors are used to plan collision-free paths, while observation sensors are used to complete the designated task.  
Different from target search problems in which the search terminates once targets are located, the goal here is to cover the entire field with observation sensors. 

We use two coordinate systems. Cartesian coordinates link to high-level objectives (e.g., to represent a point of interest in a map) and enable onboard sensor data inference (e.g., visual scene understanding). Cube coordinates are necessary to plan paths in hexagon subregions that form a hex grid. 
Thus, we use a two-layer environment map where a 2D hex grid plane is overlaid on top of a \textcolor{black}{Cartesian plane}.
\begin{figure}[h!]
\vspace{-12pt}
\centering
\includegraphics[width=0.825\linewidth]{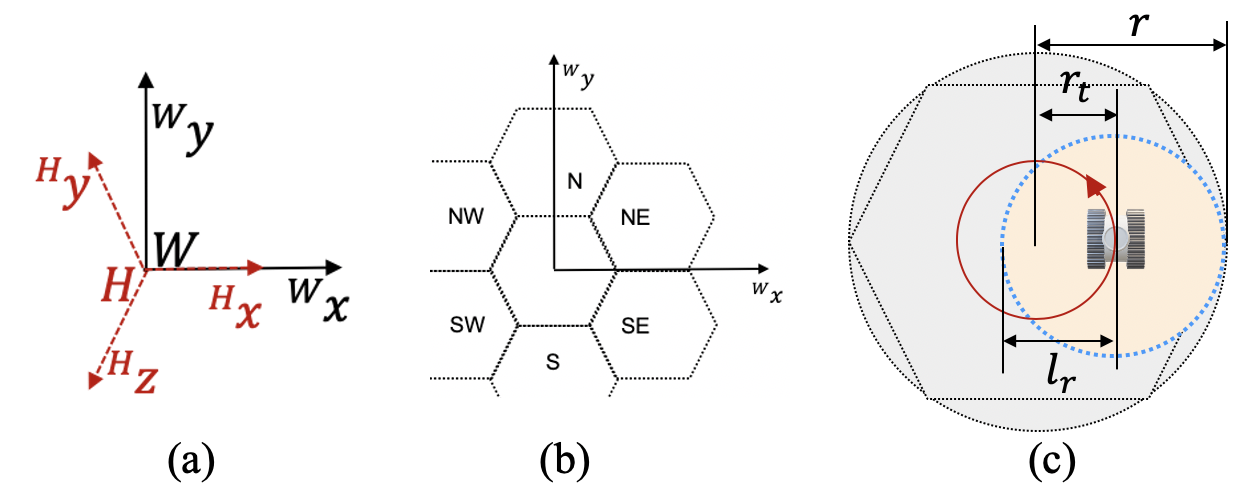}
\vspace{-6pt}
\caption{(a) Top view of the 2D hex frame $\mathit H$ (dashed red) and world frame $\mathit W$ (solid black). (b) Six adjacent cells of a subregion. (c) Covered area by the robot's observation sensors as it completes a circular path within a hex cell. 
The blue dashed circle filled with yellow depicts the sensor footprint, red solid lines represent paths, and the covered area is marked in gray.}
\label{coordinate}
\vspace{-6pt}
\end{figure}

We place a world (fixed) frame ${\mathit W}$ (Fig.~\ref{coordinate}(a) solid black lines) in \textcolor{black}{Cartesian plane}, with axes ${}^{\mathit W}x$, ${}^{\mathit W}y$. The robot is modeled in \textcolor{black}{Cartesian plane as a Dubins vehicle}, i.e. 
\begin{equation*}
{}^{\mathit W}\dot{x} = {}^{\mathit W}v\operatorname{cos}\theta\enspace\text{,}\enspace{}^{\mathit W}\dot{y} = {}^{\mathit W}v\operatorname{sin}\theta\enspace\text{,}\enspace \dot{\theta} = u\enspace \text{,}
\end{equation*}
where $(^{\mathit W}x,^{\mathit W}y)$ is the robot's position, and $\theta$ is its heading. Speed $v$ is constant, and $ u \in \{-1,0,1\}$.

We place another a frame $H$ (Fig.~\ref{coordinate}(a) dashed red lines) in hex grid plane as the frame of reference for cube coordinates. Cube coordinates correspond to three axes $({}^{\mathit H}x,{}^{\mathit H}y,{}^{\mathit H}z)$.\footnote{Cube coordinates have three axes in the 2D case.} The directions of axes are given in Fig.~\ref{coordinate}(a).\footnote{More details can be found in~\cite{Kan2019}.} The origin of frame $\mathit H$ matches frame $\mathit W$, as well as a robot's departing position. Key variables used in this paper are listed in Table~\ref{variabletable}.

\begin{table}[h]
\vspace{3pt}
\caption{List of key variables used in the paper.}\label{variabletable}
\vspace{-6pt}
\centering
\begin{tabularx}{\linewidth}{c|X}
\hline
$S_k$, $S_i$, $S_j$ & subregions  \\\hline
 $({}^{\mathit H}x,{}^{\mathit H}y,{}^{\mathit H}z)$ & cube coordinate in hex grid plane \\ \hline
 $({}^{\mathit W}x,{}^{\mathit W}y)$ & cartesian coordinate in \textcolor{black}{Cartesian plane} \\ \hline
 $r$ & hexagon grid radius \\\hline
 $r_t$ & circular path radius\\\hline
  $r_{min}$ & robot minimum turning radius\\\hline
$l_r$ & radius of observation sensor footprint \\\hline
${}^{\mathit W} \mu_k$ & starting point on circular path in $S_k$\\ \hline
$\textcolor{black}{{}^{\mathit W} \varphi_k}$ & tangent point on circular path in $S_k$ \\ \hline
$\alpha$ & angle between ${}^{\mathit W} \mu_k$ and \textcolor{black}{${}^{\mathit W} \varphi_k$}  in $S_k$\\\hline
$l$& length of straight-line path for \textit{Transitioning Mode} \\\hline
$\tau$& step index for hex grid plane planning\\\hline
\textcolor{black}{$\mathcal P_{\tau}$}& a robot's path in hex grid plane\\\hline
$\mathcal V_{\tau}$& a set of task-complete subregions\\\hline
$\mathcal E_{\tau}$&a set of explored subregions by navigation sensor\\\hline
$\Phi_{i,j}$, $\hat{\Phi}_{i,j}$& feasible path and shortest feasible path from $S_i$ to $S_j$\\\hline
\end{tabularx}
\vspace{-15pt}
\end{table}

In the hex grid plane, a hexagon subregion is $S_k$, with its cube coordinate $({}^{\mathit H}x_k,{}^{\mathit H}y_k,{}^{\mathit H}z_k)$. \footnote{We drop superscript $\mathit H$ in cube coordinates for clarity of presentation.} 
The position of the center of $S_k$ in frame $\mathit W$ is
\begin{equation*}\label{equation3}
\begin{bmatrix}
{}^{\mathit W}x
\\ 
{}^{\mathit W}y
\end{bmatrix} = \begin{bmatrix}
\frac{3}{2}r & 0 & 0\\ 
0 & \frac{\sqrt{3}}{2}r & -\frac{\sqrt{3}}{2}r
\end{bmatrix}\begin{bmatrix}
x_k & y_k & z_k
\end{bmatrix}{}^T\enspace\text{.}
\end{equation*}
%
\textcolor{black}{Each $S_k$ has six adjacent cells (Fig.~\ref{coordinate}(b)), forming a set
\begin{equation*}
\begin{medsize}
\begin{split}
    \mathcal{N}(S_k) = \{& (x_k, y_k-1, z_k+1), (x_k+1, y_k-1, z_k),\\ &(x_k+1,y_k, z_k-1), 
    (x_k, y_k+1, z_k-1),\\& (x_k-1, y_k+1, z_k), (x_k-1,y_k, z_k+1)\}.
\end{split}
\end{medsize}
\end{equation*}}
\vspace{-6pt}

Hex side length $r$ is determined based on the observation sensor footprint, taken here to be a circular disk of radius $l_r$ (Fig.~\ref{coordinate}(c)).\footnote{This is reasonable as there are a few ways to achieve a circular footprint, e.g., via rotation with gimbals or by using multiple sensors.} Radius $l_r$ depends on the selected sensor's range and is chosen by the user so that to achieve object detection of acceptable (by the user) accuracy. 
Once $l_r$ is determined, the goal is to achieve full coverage for circumscribed circle of each hexagon cell. Suppose the radius of circular paths is $r_t$, chosen such as $r_{min}\leq r_t \leq l_r$. Ideally, setting $r_t = l_r$ ensures complete coverage of the circumscribed circle of a hexagon cell with no redundancy. However, in practice we may often have $r_t < l_r$ in which case more sensor data are collected but at the expense of efficiency (defined as newly covered area per unit time). Given $l_r$ and $r_t$, then $r = l_r + r_t$.

Our method assumes the following. 1) The navigation sensor detection radius is large enough to at least detect obstacles in adjacent subregions. 2) There is sufficient battery life to achieve full coverage. 3) $r_{min}$ is reasonably small compared to the sensor footprint and obstacle density.

\vspace{-2pt}
\section{Online Hierarchical Coverage Planning}
We propose a hierarchical approach to cover an unknown environment.  
High level planning runs on the hex grid. A robot determines a sequence of subregions to visit in the next several time steps (Section~\ref{paths}). Low level planning runs on the \textcolor{black}{Cartesian plane}. A robot plans circular and straight-line paths to complete an observation task in current subregion and move to next subregion, respectively (Section~\ref{trajectory}). 

\vspace{-2pt}
\subsection{Hex Decomposition Coverage Planning}\label{paths}
We first describe our proposed Hex Decomposition Coverage Planning (HDCP) approach. HDCP works at the hex grid plane, and ensures resolution-complete coverage of hex-decomposed unknown regions. 
 In each subregion, a robot is deployed to complete the observation task, 
entering into an \textit{Observing Mode}. A subregion is \textit{visited}, if the robot has finished \textit{Observing Mode} in this subregion. Re-entering a visited subregion will not trigger another \textit{Observing Mode}. A subregion is \textit{explored} if it has been covered by navigation sensors, i.e. having been marked as obstacle-free or obstacle-occupied region. Once a robot completes an observation task within one subregion, it enters into a \textit{Transitioning Mode} and moves to another subregion. Only \textit{Observing Mode} triggers observation sensors; navigation sensors collect information consistently during both modes.

\textcolor{black}{We demonstrate the detailed process for HDCP in Algorithm~\ref{al:nextsub}. Upon departure, the robot initiates \textbf{robot-centric} frames $H$ and $W$, whose origins are at robot's departure position, for high-level and low-level planning, respectively. A bounded unknown space $\mathcal{S}$ consists of unknown but finite number of subregions $S_k(x_k,y_k,z_k)$. Let $\tau$ be the step of high-level planning which records when a subregion was visited, i.e. pair $(\tau,S_k)$ represents a robot's position in hex grid plane at step $\tau$. 
A robot's path in the hex grid plane up to step $\tau$ is then defined as $\mathcal{P_\tau} = \{(t, S_k)|t \in [1,\tau], S_k \in \mathcal{S}\}$. Let $\mathcal{V}_{\tau}$ be the set containing visited subregions up to step $\tau$, i.e. unique subregions in $\mathcal{P}_\tau$. $|\mathcal{V}_{\tau}|\leq |\mathcal{P}_\tau|$, where $|\cdot|$ denotes set cardinality.} 
Let $\mathcal{E}_{\tau} = \{(S_k, u)|S_k \in \mathcal{S}, u \in \{0, 1\} \}$ be the set containing all explored subregions and their status $u$ by navigation sensors up to step ${\tau}$. $u = 0$ represents that a subregion is obstacle-free, otherwise $u = 1$. $\mathcal{E}_{\tau, u=0}$ returns all explored obstacle-free subregions, $\mathcal{E}_{\tau, u=1}$ returns obstacle-occupied, unvisitable subregions. 

\begin{algorithm}[h!]
\caption{Hex Decomposition Coverage Planning}\label{al:nextsub}
\begin{algorithmic}[1]
\Procedure{HDCP}{}
\textcolor{black}{
\State $\tau \gets 1$,
$S_i\gets (0,0,0)$, empty sets
$\mathcal{P}_{\tau}$, $\mathcal{V}_{\tau}$, $\mathcal{E}_{\tau}$, $\hat{\Phi}_{i,j}$
\While{$S_i \neq \emptyset$}
  \State Move to $S_i$ according to Eq.~\eqref{calcualtepath} and Appendix
  \State $\mathcal{P}_{\tau} \gets \mathcal{P}_{\tau}\cup ({\tau},S_i)$
  \If{$S_i \not\in \mathcal{V}_{\tau}$}
  \State Observing mode, $\mathcal{V}_{\tau} \gets \mathcal{V}_{\tau} \cup S_i$
  \EndIf
  \State Update $\mathcal{E}_{\tau}$ by navigation sensor results
  \State $S_i, \hat{\Phi}_{i,j} \gets \text{NextHex}(S_i, \mathcal{E}_{\tau}, \hat{\Phi}_{i,j})$
  \State $\tau \gets \tau + 1$
\EndWhile}
\EndProcedure
\textcolor{black}{
\Procedure{NextHex}{$S_i$, $\mathcal{E}_{\tau}$, $\hat{\Phi}_{i,j}$}
\If{$\mathcal{Q}(S_i) \neq \emptyset$}
  \State Obtain $S_j$ according to Eq.~\eqref{objective1}, $\hat{\Phi}_{i,j} \gets \emptyset$
\ElsIf{$\mathcal{Q'}(S_i) \neq \emptyset$}
  \If{$|\mathcal{E}_{\tau}| \neq |\mathcal{E}_{\tau-1}|$ \textbf{or} 
  $\hat{\Phi}_{i,j} = \emptyset$} 
  \State Obtain $S_j$ according to Eq.~\eqref{objective2}
  \State $\hat{\Phi}_{i,j} \gets \text{GetAStarPath}(S_i, S_j,\mathcal{E}_{\tau})$
  \Else\ $S_j\gets \hat{\Phi}_{i,j}[1]$, $\hat{\Phi}_{i,j} \gets \hat{\Phi}_{i,j}[2:end]$
  \EndIf
\Else\ $S_j \gets \emptyset$, $\hat{\Phi}_{i,j} \gets \emptyset$
\EndIf
\State \Return $S_j$, $\hat{\Phi}_{i,j}$
\EndProcedure}
\end{algorithmic}
\end{algorithm}

\begin{figure}[t]
\vspace{3pt}
\centering
\includegraphics[width=0.8\linewidth]{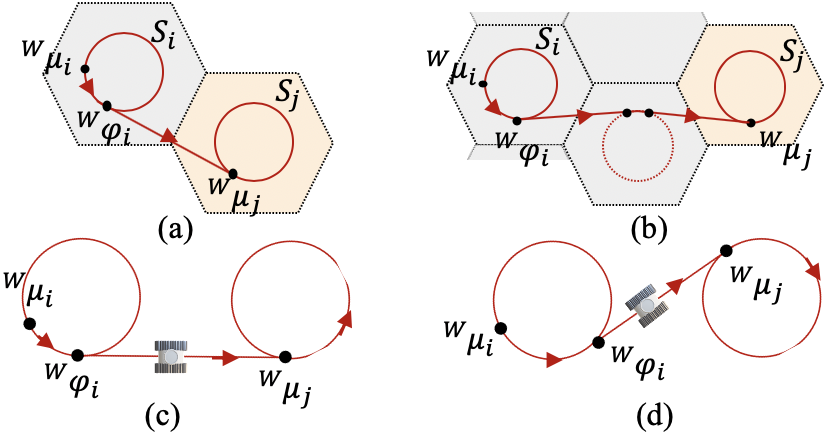}
\vspace{-12pt}
\caption{\textcolor{black}{(a) Movement to an adjacent subregion, (b) movement to a non-adjacent subregion}, (c) outer tangent line path, and (d) inner tangent line path.}
\label{movementtr}
\vspace{-15pt}
\end{figure}

\textcolor{black}{Suppose a robot finishes its observation task within subregion $S_i$ at step ${\tau}$, and needs to determine the next subregion $S_j$ to visit at step ${\tau}+1$ (line 10 of Algorithm~\ref{al:nextsub}). To decrease the number of repeatedly visited subregions, an unvisited subregion 
is preferred. 
To minimize the total travel distance, a robot prefers one of its adjacent subregions before moving to subregions further away (Fig.~\ref{movementtr}(a)). Let $\mathcal{C}_\tau = \mathcal{V}_{\tau} \cup \mathcal{E}_{\tau,u=1}$ be the set of all ``undesired'' choices of $S_j$, i.e. either already visited or obstacle-occupied. Then, set $\mathcal{Q}_{\tau}(S_i) = \{S_j^* \in \mathcal{N}(S_i) \enspace|\enspace S_j^* \not\in \mathcal{C}_\tau\}, 0 \leq |\mathcal{Q}_{\tau}(S_i)| \leq 6$ contains candidates of $S_j$, denoted as $S_j^*$, which are unvisited, obstacle-free subregions adjacent to $S_i$.}

\textcolor{black}{As we seek to complete tasks for the entire free space within the unknown region efficiently, it is undesirable to leave any isolated subregion unvisited. The cost of coming back to ``fill a hole'' later can be avoided by finishing all nearby areas first before moving away. Let function $f(\cdot)$ calculate the number of visited or obstacle-occupied neighbors of a candidate $S_j^*$ as $f(S_j^*) = |\mathcal{N}(S_j^*) \mathbin{\scalebox{1.5}{\ensuremath{\cap}}} \mathcal{C}_\tau|$. $f(S_j^*) =6$ indicates that all adjacent subregions of candidate $S_j^*$ are either visited or obstacle-occupied, which makes this $S_j^*$ a ``hole'' and hence should be prioritized to visit.}

\textcolor{black}{When $|\mathcal{Q}_{\tau}(S_i)| > 0$, there exists at least one candidate $S_j^*$ that is adjacent to $S_i$. Under this condition, among all candidates $S_j^*$, the next subregion that a robot prioritizes to visit, $S_j$, is determined as 
\begin{equation}\label{objective1}
\vspace{-3pt}
\begin{aligned}
\arg\max_{S_j^*} \quad & f(S_j^*) \\
\vspace{-3pt}
\textrm{s.t.} \quad & S_j^* \in \mathcal{Q}_{\tau}(S_i) \enspace\text{,}\enspace |\mathcal{Q}_{\tau}(S_i)| >0 \enspace\text{.}
\end{aligned}
\vspace{-2pt}
\end{equation}
}

\textcolor{black}{Thanks to~\eqref{objective1}, among all unvisited adjacent subregions the robot will choose the one with the maximum visited/obstacle-occupied neighbors as $S_j$ (line 16 of Algorithm~\ref{al:nextsub}). This strategy ensures that the robot does not leave any isolated unvisited subregions, to avoid the need to return to this area.}

\textcolor{black}{If $|\mathcal{Q}_{\tau}(S_i)| = 0$, the task has been completed for all adjacent subregions of $S_i$. Under this condition, the robot will move to a subregion $S_j$ that is nonadjacent to $S_i$ (Fig.~\ref{movementtr}(b)). 
Let $\mathcal{Q}_{\tau}(S_i)'= \{S_j^* \in \mathcal{V}_{\tau} \enspace|\enspace \mathcal{N}(S_j^*) \setminus \mathcal{C}_{\tau}\neq \emptyset\}$ be the set that contains all candidates $S_j^*$, which are visited subregions with unvisited obstacle-free neighbors. If $\mathcal{Q}_{\tau}(S_i)'$ is not empty, the robot selects $S_j$ from all candidates $S_j^*$ as
\begin{equation}\label{objective2}
\vspace{-3pt}
\begin{aligned}
\arg\max_{S_j^*} \quad & {t}\\
\vspace{-3pt}
\textrm{s.t.} \quad & (t,S_j^*) \in \mathcal{P}_{\tau}\enspace\text{,}\enspace S_j^* \in \mathcal{Q}_{\tau}(S_i)'\enspace\text{,}\enspace t\in[1,\tau] \enspace\text{.}
\end{aligned}
\vspace{-2pt}
\end{equation}
}

\textcolor{black}{
Per~\eqref{objective2}, the robot revisits a nearest-in-the-past visited subregion $S_j$ which has unvisited obstacle-free adjacent subregions (line 19 of Algorithm~\ref{al:nextsub}). Once $S_j$ is chosen, we seek a path in hex grid, which contains a sequence of subregions on the way to move to $S_j$.  In order to determine an optimal feasible path, we need some definitions.}

\begin{mydef}\label{feasible_path}
\textcolor{black}{A \textit{feasible path} between obstacle-free subregions $S_i$ and $S_j$ in hex grid plane, denoted as $\Phi(S_i,S_j)$,\footnote{$\Phi(S_i,S_j)$ is shortened as $\Phi_{i,j}$ in the following.} is a list of ordered obstacle-free subregions\footnote{The first subregion on the ordered list is adjacent to $S_i$, while the last subregion is $S_j$.} such that if only the movement to the adjacent subregion is allowed for each step, a robot starting from $S_i$ can reach $S_j$ in finite steps.}
\end{mydef}

\begin{mydef}\label{shortest_feasible_path}
\textcolor{black}{A \textit{shortest feasible path}, denoted as $\hat{\Phi}_{i,j}$, is the path among all feasible paths $\Phi_{i,j}$ which contains the least number of subregions.}
\end{mydef}

\begin{mydef}\label{visitable}
A subregion is \textit{visitable} if and only if
1) this subregion is obstacle-free, and 
2) there exists one (shortest) feasible path from the departure position to this subregion.
\end{mydef}

Each subregion in the \textit{feasible path} and \textit{shortest feasible path} is an adjacent subregion of its preceding and following subregions in the ordered list. The \textit{feasible path} and \textit{shortest feasible path} are not unique between two subregions. \textcolor{black}{Moreover, from the current position $S_i$, there must exist at least one \textit{feasible path} to any subregion on path $\mathcal{P}_\tau$, which is a subset of path $\mathcal{P}_\tau$.}

\textcolor{black}{A \textit{shortest feasible path} $\hat{\Phi}_{i,j}$ can be obtained according to function \textit{GetAStarPath} (line 20 of Algorithm~\ref{al:nextsub}). Given all explored subregions $\mathcal{E}_{\tau}$, \textit{GetAStarPath} applies the $A^*$ algorithm~\cite{Hart1968} with distance cost $q$ in 2D hex grid
 \begin{equation}\label{distance}
\vspace{-7pt}
    q(S_k,S_{k'}) = (|x_k-x_{k'}| + |y_k-y_{k'}| + |z_k-z_{k'}|)/2\enspace.
 \vspace{-7pt}   
 \end{equation} }

\textcolor{black}{While following $\hat{\Phi}_{i,j}$ (line 21 of Algorithm~\ref{al:nextsub}), more subregions are explored by the navigation sensor. If $\mathcal{E}_{\tau}$ is updated, $\hat{\Phi}_{i,j}$ needs to be updated accordingly (lines 18-20 of Algorithm~\ref{al:nextsub}). Note that $\hat{\Phi}_{i,j}$ may contain 1) unvisited subregions, and 2) subregions with unvisited obstacles-free neighbors. If at any step, an unvisited subregion is encountered, the robot enters \textit{Observing Mode} to complete the task for the subregion (lines 6-8 of Algorithm~\ref{al:nextsub}), then switches to \textit{Transitioning Mode}. If a subregion with unvisited obstacles-free neighbors is encountered, we discard $\hat{\Phi}_{i,j}$ and move to an adjacent unvisited subregion (line 16 of Algorithm~\ref{al:nextsub}).} 


\textcolor{black}{When $\mathcal{Q}_\tau(S_i) = \mathcal{Q'}_\tau(S_i) = \emptyset$, coverage is complete. HDCP establishes that in a bounded environment, for given discretization resolution, the coverage process terminates in finite time, leaving no unvisited area that is visitable according to Definition~\ref{visitable}. Coverage using HDCP is complete to the resolution of the smallest allowed hexagon cell, which can be referred as resolution-completeness~\cite{latombe2012robot}.}

\noindent  \textcolor{black}{\textbf{Lemma 1}
\textit{In an unknown, bounded environment, online coverage using HDCP (Algorithm~\ref{al:nextsub})  is resolution-complete.}}


\noindent \textbf{Proof of Lemma 1}
We prove Lemma 1 by contradiction. Assume the coverage terminates according to Algorithm~\ref{al:nextsub} and there still exists a \textit{visitable}, yet unvisited, subregion $S_q$. Suppose the departure position of the coverage is $S_p$. 
\textcolor{black}{Since $S_q$ is \textit{visitable}, there must exist a \textit{shortest feasible path} $\hat{\Phi}_{p,q}$ according to Definition~\ref{visitable}. Denote the $m$th subregion on the \textit{shortest feasible path} as $\hat{\Phi}_{p,q}[m]$.}

\textcolor{black}{
According to Definition~\ref{feasible_path}, $\hat{\Phi}_{p,q}[1]$ must be an adjacent subregion to $S_p$, $\hat{\Phi}_{p,q}[1] \in \mathcal{N}(S_p)$. 
Since the progress has terminated, $\mathcal{Q'}_{\tau}(S_p) = \emptyset$.
Recalling the definition of $\mathcal{Q'}_{\tau}$, we can deduce~\eqref{proof3} from~\eqref{proof1} and \eqref{proof2}.
%
\begin{equation}\label{proof1}
\hat{\Phi}_{p,q}[1] \in \mathcal{N}(S_p).
\end{equation} 
\begin{equation}\label{proof2}
\mathcal{Q'}_{\tau}(S_p) = \emptyset \Rightarrow \mathcal{N}(S_p) \subset \mathcal C_{\tau}.
\end{equation}
\begin{equation}\label{proof3}
\hat{\Phi}_{p,q}[1] \in \mathcal{N}(S_p),
\mathcal{N}(S_p) \subset \mathcal C_{\tau} \Rightarrow 
\hat{\Phi}_{p,q}[1]\in \mathcal C_{\tau}.
\end{equation}
\vspace{-12pt}
}

\textcolor{black}{Similarly, we have $\hat{\Phi}_{p,q}[m]\in \mathcal C_{\tau}$ for $m = 2,\cdots,|\hat{\Phi}_{p,q}|$. Therefore,  $S_q = \hat{\Phi}_{p,q}[|\hat{\Phi}_{p,q}|] \in  \mathcal{C}_{\tau} $. Hence, $S_q$ is either not \textit{visitable} or already visited, which leads to the contradiction.}

\begin{figure*}[]
\centering
\includegraphics[width=0.96\linewidth]{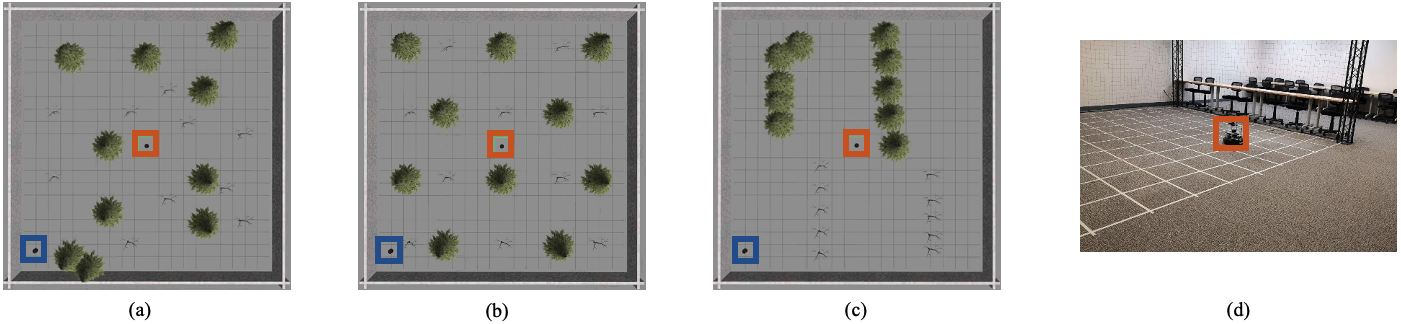}\
\vspace{-9pt}
\caption{In simulation, (a) random, (b) uniform, and (c) in-row environments. 
(d) Experimental environment.
}
\label{combined_fig}
\vspace{-3pt}
\end{figure*}

\subsection{Dubins Path Planning}\label{trajectory}
Next, we discuss Dubins Path Planning for \textit{Observing Mode} and \textit{Transitioning Mode}. 
For a robot currently in $S_i$, it first enters into \textit{Observing Mode}. For an observation task, it is required that the observation sensor covers the entire subregion, which leads to a circular path. Let ${}^{\mathit W} \mu_i$ be the starting point on the circular path. The robot follows a full circle, $ (C_{2\pi})_i$, and then returns to ${}^{\mathit W} \mu_i$. Once arriving at ${}^{\mathit W} \mu_i$, the robot enters into \textit{Transitioning Mode}, in which the robot aims to move to the starting point ${}^{\mathit W} \mu_j$ for the next subregion $S_j$. Constrained by the vehicle model, feasible paths always consist of arcs and straight lines, without turns sharper than robot's minimum turning capability. 

Figures~\ref{movementtr}(c), (d) show the planned path, which comprises an arc of angle $\alpha$ and a straight line of length $l$. The straight line is chosen to be tangent to both circular paths in $S_i$ and $S_j$. Let the tangent point in $S_i$ be \textcolor{black}{${}^{\mathit W} \varphi_i$}, the corresponding tangent point in $S_j$ will become the starting point ${}^{\mathit W} \mu_j$. Among outer tangent points (Fig.~\ref{movementtr}(c)) and inner tangent points (Fig.~\ref{movementtr}(d)), following the robot's current moving direction, we select the one that is the closest to ${}^{\mathit W} \mu_i$ along the circular path.  
Using geometry (closed-form solutions for tangent points are given in the Appendix), $\alpha$ and $l$ are 
%
\begin{equation}\label{calcualtepath}
\begin{split}
\alpha &= \operatorname{cos}^{-1}(\textcolor{black}{1}-\|{}^{\mathit W} \mu_i - \textcolor{black}{{}^{\mathit W} \varphi_i}\|^2/2r_t^2) \enspace\text{,}\enspace\\
 l &= \|{}^{\mathit W} \mu_i - \textcolor{black}{{}^{\mathit W} \varphi_i}\|\enspace\text{.}
\end{split}
\end{equation}

The combined path for \textit{Observing} and \textit{Transitioning Mode} between $S_i$ and $S_j$ is $( C_{2\pi})_i (C_{\alpha} L_{l})_{i\rightarrow j}$. By utilizing tangent points to switch among arcs and straight line paths, generated combined paths are smooth.

\subsection{Trading-off Exploration Coverage and Speed}
In some scenarios such as waypoint coverage~\cite{Karapetyan2019} and water sample collection~\cite{Manjanna2018}, full coverage within subregions does not need to be enforced. In this case, the exploration process of unknown environments can be accelerated by combining \textit{Observing} and \textit{Transitioning Mode}, which is referred to as HDCP-E. In HDCP-E, full circle trajectory $(C_{2\pi})_i$ is removed from planned paths to accelerate exploration of more hexagon subregions. At the same time, observation sensors are enabled throughout the process. The modified paths become $(C_{\alpha} L_{l})_{i\rightarrow j}$.  Hence, HDCP-E inherits the advantage of smooth and continuous paths from HDCP.

The two variants (HDCP and HDCP-E) reveal the trade-off between exploitation and exploration. HDCP guarantees resolution-complete coverage \textcolor{black}{of unknown yet bounded environments} by 1) visiting all subregions, and 2) achieving full coverage within each subregion, whereas HDCP-E enables fast and complete exploration in terms of visiting all subregions. Importantly, the next-subregion-selection strategy (Algorithm~\ref{al:nextsub}) 
can also be applied to other cell-decomposition-based coverage algorithms to obtain the order of visiting subregions. 

%
\begin{table*}[]
\centering
\caption{Results for \textcolor{black}{percentage of covered area}, runtime and exploration speed \textcolor{black}{over 10 simulation trials}}\label{resulttable}
\vspace{-6pt}
\begin{tabular}{c|llc|llc|llc}
\hline
\multirow{2}{*}{\begin{tabular}[c]{@{}c@{}}\textbf{depart at} \\ \textbf{center}\end{tabular}} & \multicolumn{3}{c|}{\textbf{random}} & \multicolumn{3}{c|}{\textbf{uniform}} & \multicolumn{3}{c}{\textbf{in-row}} \\ \cline{2-10} 
 & \multicolumn{1}{c|}{\textcolor{black}{area (\%)}} & \multicolumn{1}{c|}{runtime ($s$)} & \textcolor{black}{avg ($m^2/s$)} & \multicolumn{1}{c|}{\textcolor{black}{area (\%)}} & \multicolumn{1}{c|}{runtime ($s$)} & \textcolor{black}{avg ($m^2/s$)}  & \multicolumn{1}{c|}{\textcolor{black}{area (\%)}} & \multicolumn{1}{c|}{runtime ($s$)} & \textcolor{black}{avg ($m^2/s$)} \\ \hline
HDCP & \textcolor{black}{\textbf{83.2}$\pm$0.4} & \textcolor{black}{701.2$\pm$12.6} & \textcolor{black}{0.47} & \textcolor{black}{\textbf{80.3}$\pm$0.7} & \textcolor{black}{689.7$\pm$10.3} & \textcolor{black}{0.46} & \textcolor{black}{\textbf{79.1}$\pm$1.1} & \textcolor{black}{700.3$\pm$11.7} & \textcolor{black}{0.44} \\ \hline
HDCP-E & \textcolor{black}{68.1$\pm$0.4} & \textcolor{black}{337.3$\pm$8.0} & \textcolor{black}{\textbf{0.79}} & \textcolor{black}{68.3$\pm$0.6} & \textcolor{black}{342.7$\pm$4.4} & \textcolor{black}{\textbf{0.78}} & \textcolor{black}{66.8$\pm$0.8} & \textcolor{black}{352.4$\pm$11.3} &  \textcolor{black}{\textbf{0.75}} \\ \hline
STC & \textcolor{black}{52.7$\pm$1.4} & \textcolor{black}{561.7$\pm$12.1} & \textcolor{black}{0.37} & \textcolor{black}{58.6$\pm$0.1} & \textcolor{black}{627.6$\pm$2.6} & \textcolor{black}{0.37} & \textcolor{black}{13.5$\pm$0.1} & \textcolor{black}{144.2$\pm$0.4} & \textcolor{black}{0.37} \\ \hline
M-HDE & \textcolor{black}{70.4$\pm$1.3} & \textcolor{black}{450.0$\pm$10.6} & \textcolor{black}{0.62} & \textcolor{black}{71.8$\pm$1.7} & \textcolor{black}{485.1$\pm$15.4} & \textcolor{black}{0.58} & \textcolor{black}{68.3$\pm$2.4} & \textcolor{black}{463.6$\pm$12.2} & \textcolor{black}{0.58} \\ \hline
BA${}^*$ & \textcolor{black}{75.7$\pm$1.1} & \textcolor{black}{629.4$\pm$18.5} & \textcolor{black}{0.46} & \textcolor{black}{73.6$\pm$0.5} & \textcolor{black}{579.5$\pm$4.5} & \textcolor{black}{0.48} & \textcolor{black}{75.1$\pm$1.7}  & \textcolor{black}{635.1$\pm$7.3} & \textcolor{black}{0.45} \\ \hline
\multirow{2}{*}{\begin{tabular}[c]{@{}c@{}}\textbf{depart at}\\ \textbf{lower-left}\end{tabular}} & \multicolumn{3}{c|}{\textbf{random}} & \multicolumn{3}{c|}{\textbf{uniform}} & \multicolumn{3}{c}{\textbf{in-row}} \\ \cline{2-10} 
 & \multicolumn{1}{c|}{\textcolor{black}{area (\%)}} & \multicolumn{1}{c|}{runtime ($s$)} & \textcolor{black}{avg ($m^2/s$)} & \multicolumn{1}{c|}{\textcolor{black}{area (\%)}} & \multicolumn{1}{c|}{runtime ($s$)} & \textcolor{black}{avg ($m^2/s$)} & \multicolumn{1}{c|}{\textcolor{black}{area (\%)}} & \multicolumn{1}{c|}{runtime ($s$)} & \textcolor{black}{avg ($m^2/s$)} \\ \hline
HDCP & \textcolor{black}{\textbf{83.2}$\pm$0.4} & \textcolor{black}{727.6$\pm$6.2} & \textcolor{black}{0.45} &\textcolor{black}{\textbf{85.7}$\pm$1.3} & \textcolor{black}{740.4$\pm$11.3} & \textcolor{black}{0.46} & \textcolor{black}{\textbf{81.3}$\pm$1.4} & \textcolor{black}{699.2$\pm$20.6} & \textcolor{black}{0.45} \\ \hline
HDCP-E & \textcolor{black}{70.7$\pm$1.1} & \textcolor{black}{369.3$\pm$10.0} & \textcolor{black}{\textbf{0.75}} & \textcolor{black}{67.8$\pm$1.1} & \textcolor{black}{338.9$\pm$9.2} & \textcolor{black}{\textbf{0.79}} & \textcolor{black}{69.2$\pm$1.5} & \textcolor{black}{357.3$\pm$10.9} & \textcolor{black}{\textbf{0.76}} \\ \hline
STC & \textcolor{black}{65.8$\pm$3.7} & \textcolor{black}{622.8$\pm$19.1} & \textcolor{black}{0.42} & \textcolor{black}{59.8$\pm$1.9} & \textcolor{black}{536.5$\pm$29.6} & \textcolor{black}{0.44} & \textcolor{black}{61.0$\pm$1.6} & \textcolor{black}{502.4$\pm$26.3} & \textcolor{black}{0.48} \\ \hline
M-HDE & \textcolor{black}{74.0$\pm$1.6} & \textcolor{black}{496.9$\pm$10.6} & \textcolor{black}{0.59} & \textcolor{black}{75.5$\pm$1.8} & \textcolor{black}{510.8$\pm$22.6} & \textcolor{black}{0.58} & \textcolor{black}{69.8$\pm$1.9} & \textcolor{black}{468.3$\pm$20.4} & \textcolor{black}{0.59} \\ \hline
BA${}^*$ & \textcolor{black}{80.7$\pm$0.7} & \textcolor{black}{661.1$\pm$21.2} & \textcolor{black}{0.46} & \textcolor{black}{79.0$\pm$1.3} & \textcolor{black}{522.8$\pm$30.0} & \textcolor{black}{0.57} & \textcolor{black}{79.2$\pm$4.3} & \textcolor{black}{666.2$\pm$52.5} & \textcolor{black}{0.47} \\ \hline
\end{tabular}
\vspace{-12pt}
\end{table*}

\section{Experiments, Results, and Discussion}
Our proposed variants are evaluated in Gazebo simulation and experimentally with a non-holonomic wheeled robot. Their performance is compared against 1) online BA${}^*$~\cite{Hoang2013}, 2) online STC~\cite{Gabriely2001}, and 3) M-HDE~\cite{Kan2019} in terms of total coverage area and average exploration speed.

\subsection{Simulation}

\subsubsection{Simulation Setup}
Figures~\ref{combined_fig}(a)-(c) show three $20 \text{m} \times 20$m 2D simulated forest/farm-like environments in Gazebo (random, uniform, and in-row placement, respectively). Two type of trees, which are different in terms of size and shape, are used. 
Ten trees of each type are placed in each environment, hence all three environments have same amount of free space. In the random environment (Fig.~\ref{combined_fig}(a)), trees are placed randomly to represent a forest-like unstructured environment. In the uniform environment (Fig.~\ref{combined_fig}(b)), trees are arranged and lined up strictly
. In the in-row environment (Fig.~\ref{combined_fig}(c)), trees are loosely lined up, with slightly-varying spacing between them. This environment approximates a more realistic intercropping agricultural field. 
\textcolor{black}{We further consider the baseline scenario of operating in an empty environment bounded by square walls.}

We deploy the non-holonomic wheeled robot Turtlebot to cover the entire area without prior knowledge of the environment map. The robot is equipped with an RPLidar laser scanner as the navigation sensor, and an Astra Pro stereo camera as the observation sensor. 
Both perception\footnote{We use the open-source LIDAR-based obstacle detector published in https://github.com/tysik/obstacle$\_$detector.} and planning are online. \textcolor{black}{To reduce the uncertainty in obstacle detection caused by online perception and odometry drift, we run \textcolor{black}{ten trials} for every obstacle-cluttered scenario.}

Camera observations are chosen to have high accuracy with $l_r=0.5$ m. Hence, for HDCP, HDCP-E, and M-HDE, the hex subregion side length is $r = 1$\;m. The robot moves at a constant velocity of $1$\;m/s. For BA${}^*$ and STC, the square side length 
is $1$\;m;\footnote{We refer the reader to~\cite{Hoang2013,Gabriely2001} for more details.} the robot moves forward at $1$\;m/s, and decelerates to $0.3$\;m/s when taking sharp turns.\footnote{We optimized the turning speed so that no odometry drift is observed.}  

To eliminate the influence of starting position, the robot is deployed from 2 different positions: center and lower-left corner (red and blue squares in Fig.~\ref{combined_fig}(a)-(c)). \textcolor{black}{Note that in both simulation and real experiment, robot-centric hex (for HDCP, HDCP-E, and M-HDE) and square (for STC and BA${}^*$) grids originate at robot's departure position. In a robot-centric grid, the number of occupied cells caused by obstacles and environment boundaries is influenced by the relative position between the object and the origin of the grid. For instance, a small obstacle can either lie within one cell, or on an edge/intersection of multiple cells after varying the grid origin. In addition, the size of cells is determined according to the robot footprint. The occupied cells near boundaries are marked as obstacles by the robot at runtime via onboard perception. We do not pre-determine near-boundary inaccessible cells for the robot prior to departure.}

\subsubsection{Results and Discussion}
\textcolor{black}{Table~\ref{resulttable} contains means and one-standard deviations for the percentage of coverage area over total free space, 
total algorithm runtime, 
and averaged exploration speed  
for each scenario over ten trials. We consider the total free space as subtracting the tree-occupied area from the total environment area, which is consistent among all scenarios.} 

\begin{figure}[ht]
\centering
\includegraphics[width=0.99\linewidth]{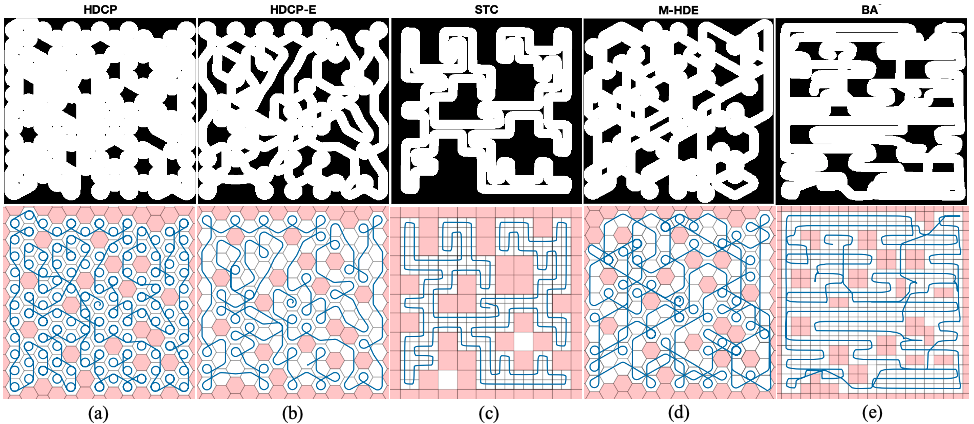}
\vspace{-10pt}
\caption{\textcolor{black}{Covered area (top panels, in white), detected obstacles (bottom panels, red cells), and corresponding robot paths (blue curves) for (a) HDCP, (b) HDCP-E, (c) M-HDE, (d) STC, and (e) BA${}^*$ in the random map when the robot departs at center position.}}
\label{combined_traj}
\vspace{-6pt}
\end{figure}

\textcolor{black}{Results suggest that, regardless of departing position, HDCP covers the most area in all evaluated, obstacle-cluttered environments. While the uniform and in-row environments bounded by square walls are more regular and structured, HDCP still outperforms the other evaluated methods that use a square grid discretization. 
Our findings demonstrate the advantage of hexagon decomposition, where we mark obstacles as more ``round-like'' hexagons instead of square cells in obstacle-cluttered environments.}

In terms of exploration speed, HDCP-E covers almost twice as fast as STC and BA${}^*$ in all environments from both departure positions. The results suggest that the strategy of HDCP and HDCP-E for selecting the next subregion is efficient in terms of exploring more unknown space. However, in HDCP, more area is covered when following the full-circle at the expense of exploration speed. M-HDE from our previous work~\cite{Kan2019} achieves the second fastest exploration, in which the lack of inner-tangent straight-line transitioning causes longer paths.\footnote{Note that we modify M-HDE by replacing the original path finding strategy in~\cite{Kan2019} with A* to achieve better performance.} Hence, in scenarios when the speed of exploring more unknown space is the main concern, HDCP-E can be used to achieve fast exploration. 

It is worth noticing that STC is the approach most sensitive to the environment and departure position. Especially in the in-row environment, STC is unable to cover the entire space when departing from center. This is because STC assumes all visitable space has to have a width of at least four times of the sensor footprint radius, to ensure repetition-free paths~\cite{Gabriely2001}. If the assumption is not satisfied, the area will be marked as obstacle, even if the width of the area is wider than the robot's own width and can thus be visited. On the other hand, for HDCP, changing departure position and environment can cause at most \textcolor{black}{$6\%$} coverage percentage loss. 

Evident in Fig.~\ref{combined_traj}, our proposed method generates smooth paths for robots with non-holonomic constraints. The robot moves at constant speed throughout the process, which enables better path following and sensor stability. In contrast, small areas can remain uncovered in both BA$^{}*$ and STC when the robot fails to follow planned paths exactly due to abrupt deceleration before turning. The advantage of smooth paths is more obvious in cluttered environments, in which case turning maneuvers are required more frequently. 

\textcolor{black}{We also evaluate the algorithm in an square environment without obstacles, which is believed to be most suitable for lawnmower-like methods such as BA${}^*$ and STC. In the empty environment, BA${}^*$ ($0.75\ m^2/s$) achieves comparable coverage speed as HDCP-E ($0.82\ m^2/s$). This is because when applying lawnmower-like methods in the empty environment, the robot follows straight-line paths from one side to another, which requires minimum number of turns. In addition, BA${}^*$ ($92.1\%$) covers slightly more area than HDCP ($91.4\%$). This is because square cells describe the square environment better when no obstacle exists. However, in obstacle-cluttered environments, the existence of unexpected obstacles force the robots to take frequent, possibly sharp turns. Taking sharp turns requires deceleration and acceleration for lawnmower-like methods, whereas HDCP and HDCP-E allow robots to operate at constant velocity. Moreover, as discussed above, the advantage of describing obstacles with hex cells is more obvious in obstacle-cluttered environments than empty environments. Overall, the proposed HDCP and HDCP-E work well in unknown and irregularly-shaped obstacle-cluttered environments that are bounded. }

\subsection{Experiments}
We also evaluate the performance of all algorithms in a $10\text{m} \times 8\text{m}$ indoor space (Fig.~\ref{combined_fig}(d)) with a real Turtlebot robot configured as in the simulation. The environment contains a truss, desks, and chairs as obstacles.  
For HDCP, HDCP-E, and M-HDE, we have $l_r=0.4$\;m and 
$r = 0.8$\;m; the robot moves at constant velocity of $0.3$\;m/s. For BA${}^*$ and STC, we have 
side length of $0.8$\;m, and the robot moves forward at $0.3$\;m/s, and decelerates to $0.1$\;m/s in turning. 

\begin{table}[h]
\vspace{-3pt}
\caption{\textcolor{black}{Percentage of covered area, runtime and exploration speed in experiments}}\label{experimentresult}
\vspace{-9pt}
\centering
\begin{tabular}{c|c|c|c|c|c}
\hline
 & HDCP & HDCP-E & STC & M-HDE & BA${}^*$ \\ \hline
\textcolor{black}{area (\%)} & \textcolor{black}{\textbf{76.5}} & \textcolor{black}{62.3} & \textcolor{black}{65.6} & \textcolor{black}{73.5} & \textcolor{black}{75.8} \\ \hline
runtime ($s$) & 367.0 & 187.5 & 268.5 & 259.0 & 306.5 \\ \hline
\textcolor{black}{avg ($m^2/s$)} & \textcolor{black}{0.10} & \textcolor{black}{\textbf{0.16}} & \textcolor{black}{0.12} & \textcolor{black}{0.14} & \textcolor{black}{0.12} \\ \hline
\end{tabular}
\end{table}

\begin{figure}[t]
\vspace{4pt}
\centering
\includegraphics[width=1\linewidth]{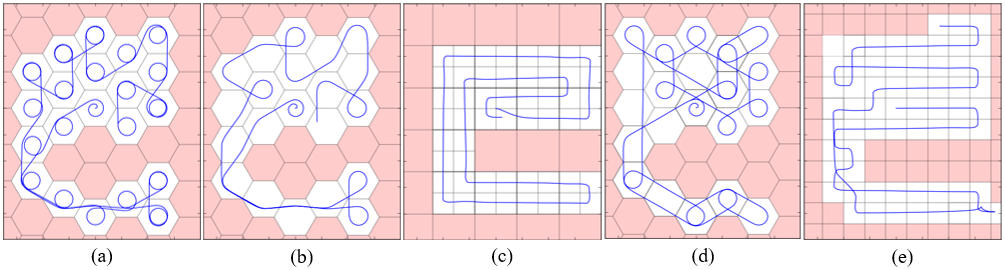}
\vspace{-15pt}
\caption{Paths for (a) HDCP, (b) HDCP-E, (c) STC, (d) M-HDE, and (e) BA${}^*$ in hardware experiments.}
\label{traj_robot}
\vspace{-12pt}
\end{figure}

We observe that HDCP and BA${}^*$ cover similar area, which is expected \textcolor{black}{since the experimental environment is mostly empty with obstacles lined up strictly in the middle}. HDCP-E has the highest average exploration speed, $0.16\ \text{m}^2/\text{s}$, which is consistent to simulation results. Further, our approach generates smooth paths (Fig.~\ref{traj_robot}) in real time for robots to follow at constant speed, while BA${}^*$ and STC both require frequent acceleration and deceleration since the robot makes turns frequently in the small, cluttered space.

\section{Conclusions}
The paper contributes to \textit{online resolution-complete coverage planning} in unknown obstacle-cluttered environments. Research on this vein is limited when it comes to considering some form of dynamic feasibility (in this case Dubins vehicles), while ensuring resolution-complete coverage.

Results suggest that our proposed algorithm HDCP can cover more area compared to existing methods such as M-HDE, STC and BA${}^*$ in unstructured and obstacle-cluttered environments, due to the advantage of decomposing the workspace in hex cells. Its variant HDCP-E achieves the fastest exploration (covered area per unit time) in both structured and unstructured environments. Further, we show that Dubins vehicles may fail to follow frequently required sharp turns using STC and BA${}^*$, leading to uncovered areas along the search path. Our method guarantees resolution-complete coverage while considering non-holonomic constraints in the form of Dubins curves. Derived geometric closed-form solutions to determine how to move between subregions enable real-time planning. Current weaknesses of HDCP are a slight underperformance compared to BA${}^*$ in square, empty environments, and the presence of repeated arc segments on circular paths when moving to the tangent points. However, we believe that ensuring path smoothness and continuity due to this repetition outweighs the limitation.

Future work will focus on performing a complexity analysis to investigate how to speed up the methodology, and \textcolor{black}{application to real world agricultural/forest environments}.

\section*{APPENDIX}
Suppose the center position of $S_k$, $S_{k'}$ in the \textcolor{black}{Cartesian plane} are $(a_k,b_k)$, $(a_{k'},b_{k'})$, respectively. $w = (a_{k'}-a_k)^2 + (b_{k'}-b_k)^2$. Inner tangent points are
\begin{equation*}
\vspace{-2pt}
\begin{medsize}
\begin{split}
\textcolor{black}{{}^{\mathit W} \varphi_k (x)} &= \frac{2r_t^2(a_{k'}-a_k)\pm r_t(b_{k'}-b_k)\sqrt{w-(2r_t)^2}}{w}+a_k\enspace\text{,}\\
\textcolor{black}{{}^{\mathit W} \varphi_k (y)} &= \frac{2r_t^2(b_{k'}-b_k)\pm r_t(a_k-a_{k'})\sqrt{w-(2r_t)^2}}{w}+b_k\enspace\text{,}\\
{}^{\mathit W} \mu_{k'}(x) &= \frac{2r_t^2(a_k-a_{k'})\pm r_t(b_k-b_{k'})\sqrt{w-(2r_t)^2}}{w}+a_{k'}\enspace\text{,}\\
{}^{\mathit W} \mu_{k'}(y) &= \frac{2r_t^2(b_k-b_{k'})\pm r_t(a_{k'}-a_k)\sqrt{w-(2r_t)^2}}{w}+b_{k'}\enspace\text{.}
\end{split}
\label{eq5}
\end{medsize}
\vspace{-2pt}
\end{equation*}
Outer tangent points are
\begin{equation*}
\begin{medsize}
\begin{split}
\textcolor{black}{{}^{\mathit W} \varphi_k (x)} = a_k \pm r_t\frac{(b_k-b_{k'})}{\sqrt{w}}\enspace &\text{,} \enspace \textcolor{black}{{}^{\mathit W} \varphi_k (y)} = b_k \pm r_t\frac{(a_{k'}-a_k)}{\sqrt{w}}\enspace \text{,}\\
{}^{\mathit W} \mu_{k'}(x) =a_{k'} \pm r_t\frac{(b_k-b_{k'})}{\sqrt{w}}\enspace &\text{,}\enspace
{}^{\mathit W} \mu_{k'}(y) = b_{k'} \pm r_t\frac{(a_{k'}-a_k)}{\sqrt{w}}\enspace\text{.}
\end{split}
\label{eq6}
\end{medsize}
\end{equation*}

\balance
\bibliographystyle{IEEEtran}
\bibliography{IEEEabrv,IEEEexample}

\end{document}